\pdfoutput=1
\documentclass[11pt]{article}

\usepackage[final]{coling}

\usepackage{times}
\usepackage{latexsym}

\usepackage[T1]{fontenc}
\usepackage[utf8]{inputenc}

\usepackage{microtype}

\usepackage{inconsolata}

\usepackage{graphicx}
\usepackage{soul}

\title{NüshuRescue: Revitalization of the Endangered Nüshu Language with AI}

\author{
Ivory Yang$^1$ \quad 
Weicheng Ma$^2$ \quad 
Soroush Vosoughi$^3$ \\
$^{1,2,3}${Department of Computer Science, Dartmouth College} \\
{\texttt{\{Ivory.Yang.GR, Weicheng.Ma, Soroush.Vosoughi\}@dartmouth.edu}}\\
}

\begin{document}
\maketitle
\begin{abstract}
The preservation and revitalization of endangered and extinct languages is a meaningful endeavor, conserving cultural heritage while enriching fields like linguistics and anthropology. However, these languages are typically low-resource, making their reconstruction labor-intensive and costly. This challenge is exemplified by Nüshu, a rare script historically used by Yao women in China for self-expression within a patriarchal society. To address this challenge, we introduce NüshuRescue, an AI-driven framework designed to train large language models (LLMs) on endangered languages with minimal data. NüshuRescue automates evaluation and expands target corpora to accelerate linguistic revitalization. As a foundational component, we developed NCGold, a 500-sentence Nüshu-Chinese parallel corpus, the first publicly available dataset of its kind. Leveraging GPT-4-Turbo, with no prior exposure to Nüshu and only 35 short examples from NCGold, NüshuRescue achieved 48.69\% translation accuracy on 50 withheld sentences and generated NCSilver, a set of 98 newly translated modern Chinese sentences of varying lengths. In addition, we developed FastText-based and Seq2Seq models to further support research on Nüshu. NüshuRescue provides a versatile and scalable tool for the revitalization of endangered languages, minimizing the need for extensive human input. All datasets and code have been made publicly available at \url{https://github.com/ivoryayang/NushuRescue}.
\end{abstract}

\begin{figure}[t]
  \centering
  \includegraphics[width=\linewidth]{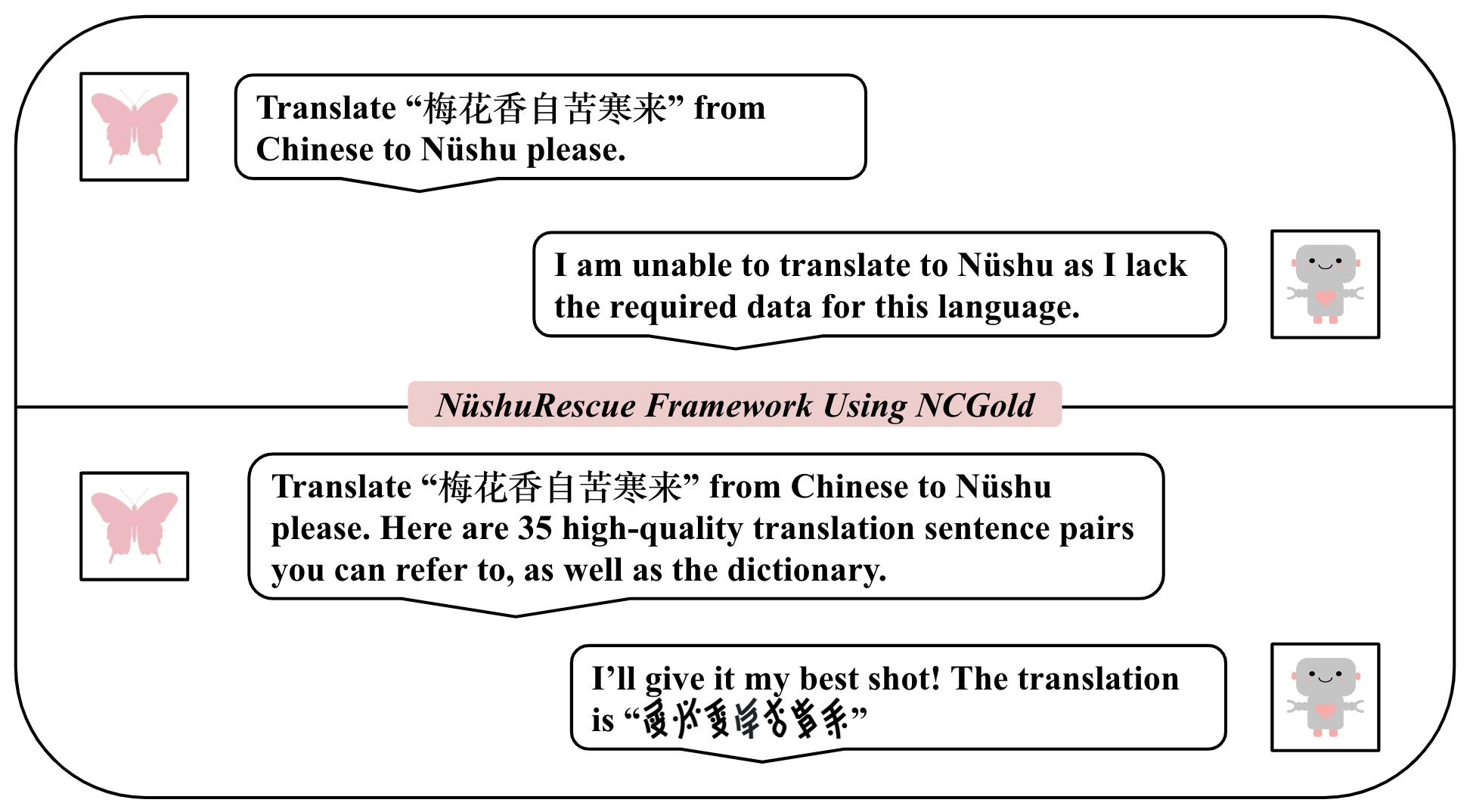}
  \caption{A simplified, stylized rendition of the NüshuRescue generation framework, using samples from NCGold.}
  \label{fig:illustration:1}
\end{figure}

\section{Introduction}
Nüshu is a unique syllabic script originating from the ethnic Yao women in Jiangyong County, Hunan Province, China \cite{broussard2008nushu, liu2014biographical}. This rare and historically significant script, often referred to as ``women's script'' \cite{wang2020nushu}, is the only written language in the world developed by and used exclusively by women \cite{ann2013sustaining}. In a past era of high female illiteracy, where women's voices were largely suppressed \cite{leung2003women}, Nüshu served as a private means of communication and form of female empowerment in a male-dominated society \cite{liu1997women}. Today, Nüshu is an endangered language with very limited surviving documentation \cite{luo2022tourism}, making it a critical focus for linguistic preservation and revitalization efforts \cite{liu2018practice}.

In the broader field of Natural Language Processing (NLP), low-resource languages like Nüshu present significant challenges, such as the scarcity of available data \cite{magueresse2020low} and the absence of established linguistic tools \cite{joshi2019unsung, krasadakis2024survey}. The preservation of such languages is not only vital for maintaining linguistic diversity \cite{gibbs2002saving, brenzinger2009documenting}, but also for safeguarding the cultural heritage \cite{romaine2007preserving} and historical narratives \cite{hale1992endangered} they encapsulate. To preserve the Nüshu language, this paper provides two main contributions: \textbf{(1) NüshuRescue:} A LLM-based data generation framework to construct or scale up text corpora for low-resource languages, using minimal human labor and without presuming any prior knowledge of the target languages. \textbf{(2.1) NCGold:} An expert-validated parallel corpus of 500 Nüshu-Chinese translation sentence pairs. \textbf{(2.2) NCSilver:} A parallel corpus of 98 Nüshu-Chinese translation sentence pairs generated through the NüshuRescue framework. Both NCGold and NCSilver are the first publicly available datasets of their kind.

Through our experiments, we find that LLMs like GPT-4-Turbo demonstrate strong language learning capabilities, even for languages with limited prior exposure, allowing the NüshuRescue framework to effectively generate higher-quality text for truly resource-scarce languages. While human expertise remains crucial for annotating seed data and ensuring translation accuracy, NüshuRescue significantly reduces the amount of human labor required compared to fully manual data annotation. As such, we believe that our NüshuRescue framework, as shown in Figure \ref{fig:illustration:1}, offers a promising solution for the preservation and revitalization of endangered languages.

\section{Related Work}
\label{sec:related}
Nüshu has garnered renewed cultural interest in recent years \cite{thurnell2022introduction}, especially after receiving international representation in the Hollywood production ``Snow Flower and the Secret Fan'' \cite{dargis2011ties}. However, existing linguistic work on Nüshu is limited; ``A Compendium of Chinese Nüshu'' \cite{zhao1992nushu_compendium}\footnote{The very first version of this book was published in 1992; we referenced the most recent e-book released in 2019.} remains the only formal print publication with expert-validated translations. \citet{chiu2012heroines} provided a new perspective in ``Heroines of Jiangyong'' by compiling Nüshu scripts alongside English translations of their Chinese counterparts. While this is one step removed from a direct Nüshu-to-English translation and lacks formal linguistic mapping, it is nonetheless valuable for introducing Nüshu to the English-speaking domain.

There have been only tentative attempts to explore Nüshu within the NLP community. \citet{tang2023ai} and \citet{sun2023ai} proposed a two-part project combining research and art, where two AI agents trained on a Chinese-Nüshu dictionary collaborate to create a standard writing system for encoding Chinese, blending linguistic analysis with artistic interpretation. However, no publicly available digital dataset for Nüshu exists, and using NLP methods to analyze its syntax and semantics remains virtually unexplored. The only significant digitalization effort to date has been the creation of a Nüshu-Chinese Unicode dictionary \cite{nushu_unicode_repo}, which remains a work in progress as it continues to be expanded and standardized.

\begin{figure}[t]
  \centering
  \includegraphics[width=\linewidth]{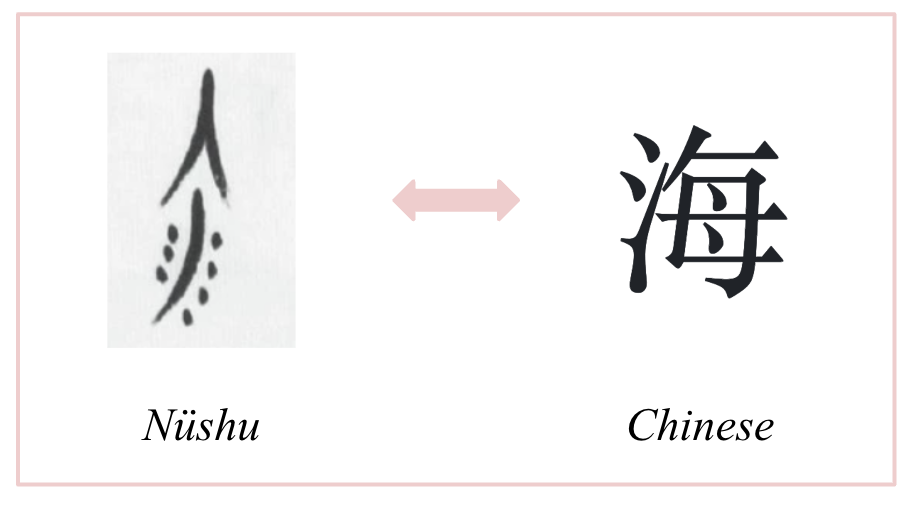}
  \caption{Comparison of mapped Chinese (logographic) and Nüshu (syllabic) characters for the word ``ocean''.}
  \label{fig:illustration:2}
\end{figure}

Efforts to document and preserve other under-resourced and endangered languages have been more thoroughly explored. A common challenge across endangered language preservation projects is the need to gather data at scale. One approach is the collection of text corpora through web crawling and data extraction, as demonstrated by \citet{de2024data} for Tetun, a language of Timor-Leste. However, this method relies on the presence of online resources, which is not the case for Nüshu. Another approach, transcription, and human annotation is more costly and labor-intensive, as seen in efforts to preserve Yoruba, a West African language \cite{fagbolu2015digital}. Given these constraints, it is crucial to combine human annotation efforts with AI and NLP tools, so as to advance Nüshu preservation in a cost-effective and scalable manner.

\section{The Nüshu Language}
Nüshu is a writing system originally created and exclusively used by ethnic Yao women in rural China \cite{bissessar2013movement}, and therefore possesses internal linguistic and literary rules difficult for those outside the community to fully grasp. Nüshu emerged independently as a phonetic writing system, borrowing sounds from spoken Chinese (particularly the local dialect) and representing them using a syllabic system \cite{zhao2014nyushu}. Nüshu is \textit{syllabic}, meaning it uses symbols to represent individual syllables in spoken language \cite{silber1995nushu}. In contrast, Chinese is \textit{logographic}, meaning it uses characters that represent words or morphemes, with each character typically corresponding to a meaning rather than a specific sound \cite{tan2001neural}. For example, Figure \ref{fig:illustration:2} illustrates the Nüshu and Chinese characters for the word ``ocean'', shown on the left and right, respectively. In Chinese, the character for ``ocean'' is logographic and conveys the concept as a whole, with the ``water radical'' (the three strokes on the left of the word) indicating its connection to water. Removing the ``water radical'' changes the character's meaning to ``each'' or ``every'', illustrating how radicals\footnote{In Chinese, radicals are the building blocks of Chinese characters, often used as a means to organize them in dictionaries. They usually represent common semantic or phonetic components.} influence the meaning of Chinese characters. In contrast, the Nüshu symbol for ``ocean'' is syllabic, representing the phonetic sounds of the spoken word rather than the concept itself.

\begin{figure}[t]
  \centering
  \includegraphics[width=\linewidth]{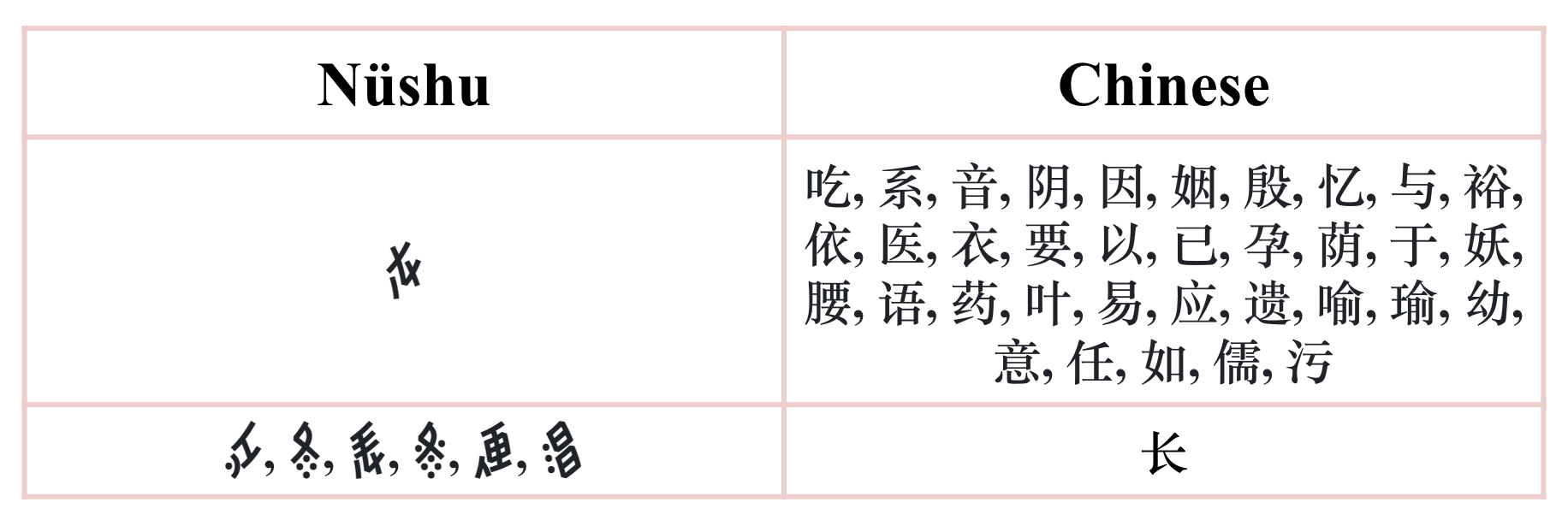}
  \caption{Nüshu-Chinese one-to-many and many-to-one mapping; in extreme cases, one Nüshu word can map up to 35 Chinese words.}
  \label{fig:illustration:3}
\end{figure}

The phonetic nature of Nüshu creates a multi-mapping challenge with Chinese, as shown in Figure \ref{fig:illustration:3}. A single Nüshu symbol can correspond to multiple Chinese characters with the same pronunciation but different meanings (one-to-many mapping). In contrast, a single Chinese character may map to several Nüshu symbols representing different phonetic variations (many-to-one mapping) \cite{moratto2022nushu}. Although an estimated 600–700 Nüshu characters exist, only 398 have been officially encoded in Unicode Noto Sans Nüshu \cite{emerson2002chinese}, leaving many symbols undocumented and creating gaps in translation efforts. Traditional methods like word alignment models \cite{och2003statistical} and statistical machine translation \cite{sharma2023machine}, which rely on large, parallel corpora, struggle to handle Nüshu's complex relationships and limited dataset. Advanced techniques are needed to accommodate Nüshu's unique structure and support corpus expansion and alignment with other languages.

\section{NüshuRescue} \label{sec:experiments}
This section details our efforts in (1) initializing NLP research on Nüshu via manual translation (Section \ref{sec:experiments:manual}) and ChatGPT-in-the-loop data annotation (Section \ref{sec:experiments:gpt}), (2) training a base language model for Nüshu (Section \ref{sec:experiments:fasttext}), and (3) enabling machine translation from Nüshu to Chinese (Section \ref{sec:experiments:n2c}) or from Nüshu to multiple other languages (Section \ref{sec:experiments:n2o}).

\subsection{Manual Data Collection} \label{sec:experiments:manual}
"A Compendium of Chinese Nüshu" \cite{zhao1992nushu_compendium} is the most comprehensive, expert-validated collection of Nüshu calligraphy available \cite{lo2013she}. It features scanned Nüshu scripts, created with paper and ink, positioned at the top of each page, with the corresponding Chinese translation at the bottom. A sample from the book is shown in Figure \ref{fig:appa} in the Appendix. While most works consist of traditional ancient Chinese poems and literature, the collection also includes personal diary entries and an impassioned soliloquy written in response to a mother's scolding, providing contextual diversity.

\begin{figure}[t]
  \centering
  \includegraphics[width=\linewidth]{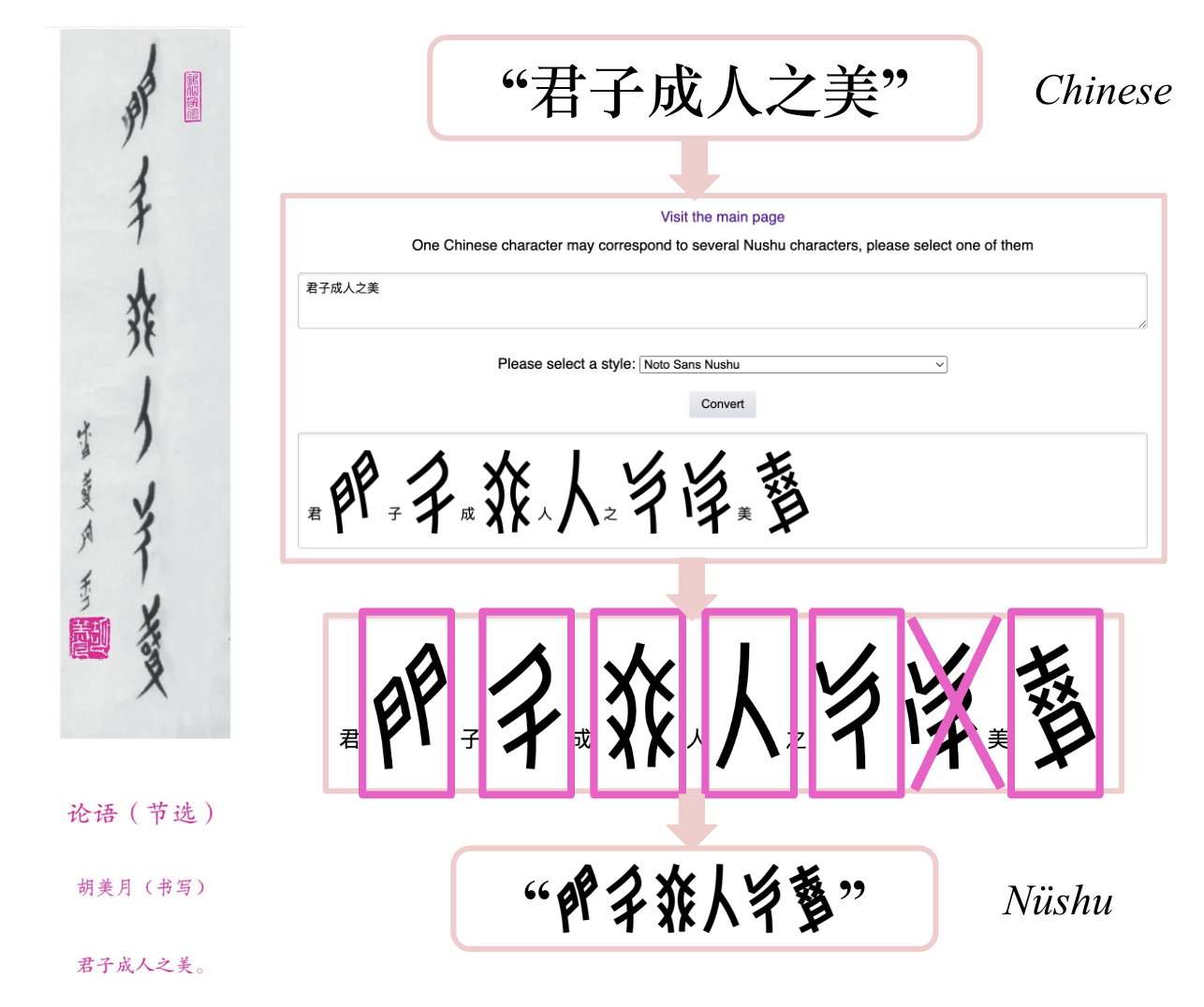}
  \caption{Manual Data Collection: Step-by-step transcription process.}
  \label{fig:illustration:4}
\end{figure}

The manual data collection process for the 500-sentence NCGold dataset was carried out with meticulous attention to detail. The annotators for the task are college-educated, bilingual in English and Chinese, and have experience in computational linguistics. The step-by-step process is illustrated in Figure \ref{fig:illustration:4}\footnote{English Translation: ``A noble man empowers the beauty within others''.}. First, each sentence of the Chinese text translations was manually typed out, with the punctuation left out. Next, the Chinese text was copied and pasted into the official online website, Converter of Calligraphy Copybook of Nüshu Standard Characters \cite{nushuscript_website}, to retrieve all possible mapped Nüshu words for each Chinese word. Finally, the annotators carefully filtered and matched the single correct Nüshu word to each corresponding Chinese word.

\begin{figure}[t]
  \centering
 \includegraphics[width=0.5\linewidth]{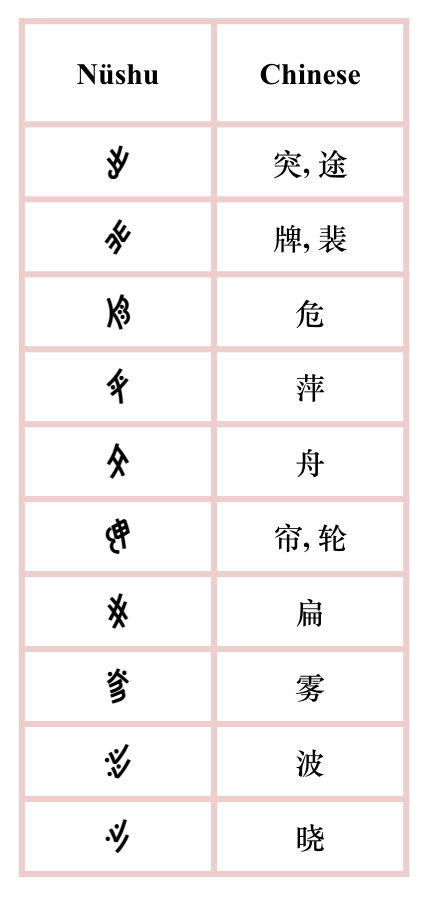}
  \caption{A small sample of newly mapped words not in the official Nüshu-Chinese dictionary; an original contribution.}
  \label{fig:chars}
\end{figure}

Of the 500 total sentences, 127 had full coverage in the official Nüshu Unicode dictionary, with every Chinese word included. For the remaining 373 sentences, while some Chinese words were covered, others were missing from the limited vocabulary of the dictionary, meaning the online converter could not provide Nüshu mappings for all terms. In these ambiguous cases, we manually consulted the Nüshu column of the official dictionary and visually identified the correct Nüshu characters from the book. As a result, NCGold not only provides the first expert-validated Nüshu-Chinese parallel corpus but also contributes newly mapped Nüshu-Chinese word pairs, a small sample of which is shown in Figure \ref{fig:chars} (Figure \ref{fig:appb} in the Appendix shows a higher resolution version).

\subsection{LLM-Based Data Augmentation} \label{sec:experiments:gpt}
After obtaining NCGold, we established a text translation framework for scaling up the dataset by leveraging LLMs. For clarity, we refer to the manually-translated dataset as \textbf{NCGold} and the subsequent LLM-generated dataset as \textbf{NCSilver}.

\begin{figure}[t]
  \centering
  \includegraphics[width=\linewidth]{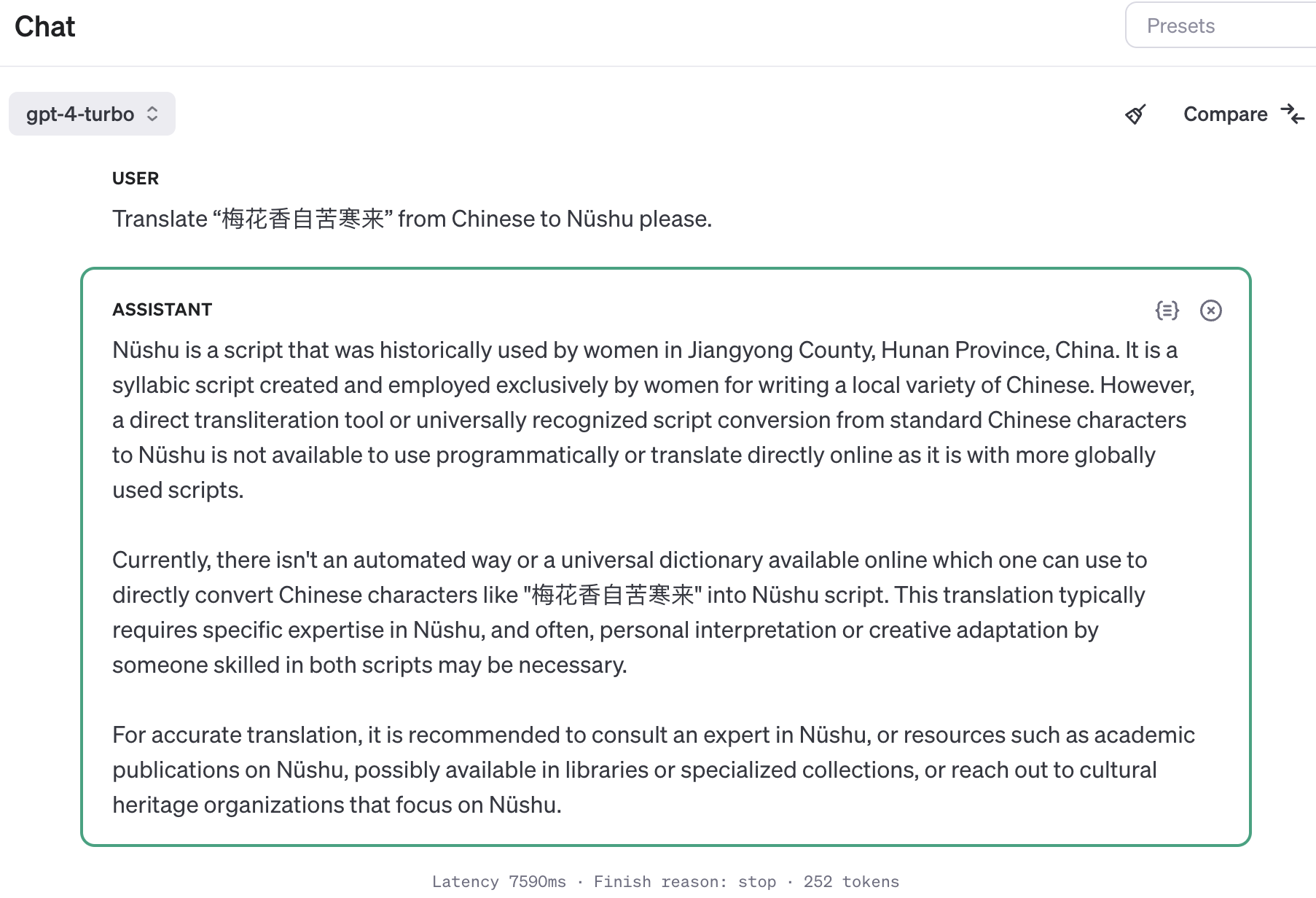}
  \caption{GPT-4-Turbo does not understand Nüshu at the beginning.}
  \label{fig:model-knowledge}
\end{figure}

One notable merit of our approach is that the LLM does not require any prior knowledge of Nüshu characters before being introduced to the provided examples, as demonstrated in Figure \ref{fig:model-knowledge} \footnote{English Translation: ``A fragrant plum blossom comes from the bitter winter cold''. This is a traditional Chinese idiom that means for good things to happen, one often needs to go through hardship.}
While we use the GPT-4-Turbo model in our experiments, our proposed framework is model-agnostic, allowing for the use of any LLM with comparable language modeling and reasoning capabilities, as illustrated in Figure \ref{fig:appc} in the Appendix.

\begin{figure}[t]
  \centering
  \includegraphics[width=\linewidth]{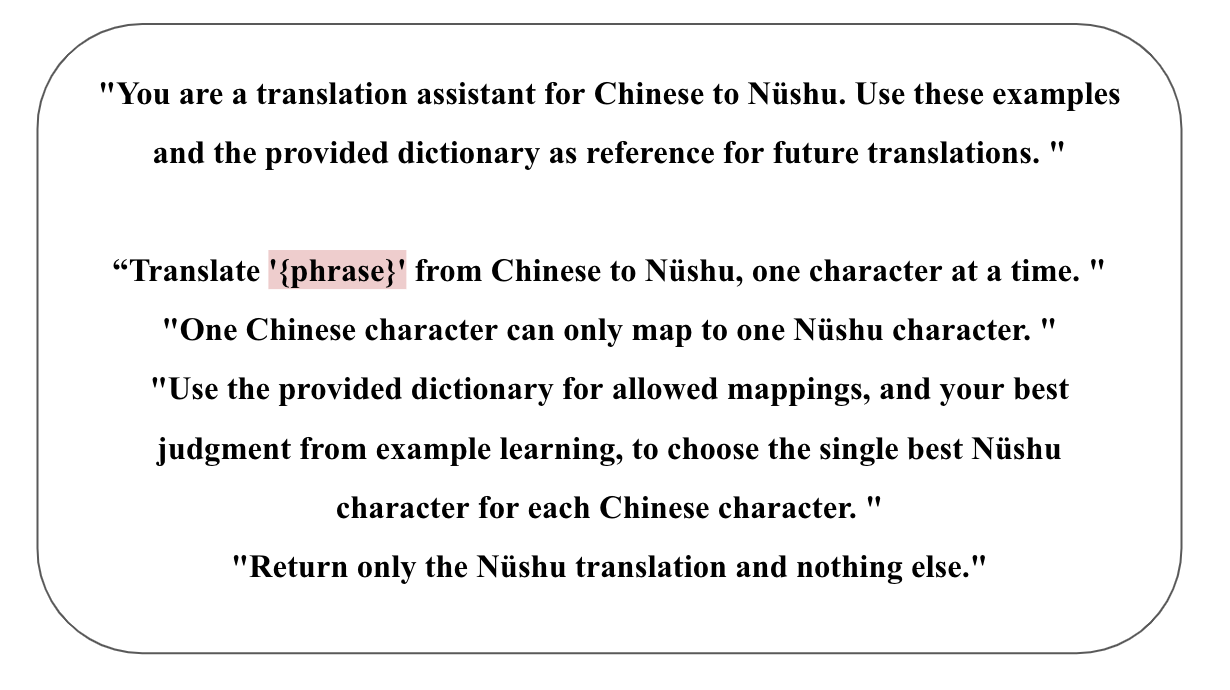}
  \caption{Prompt used on GPT-4-Turbo.}
  \label{fig:gpt-data-annotation-example}
\end{figure}

\subsubsection{Experimental Settings}
In our framework for obtaining Nüshu translations from the LLMs, we first provide translation examples from NCGold and the official Nüshu-Chinese dictionary for GPT-4-Turbo to learn from, uploading them as files to avoid excessively long prompts. The model is then prompted to generate Nüshu translations for Chinese input sentences, as shown in Figure \ref{fig:gpt-data-annotation-example}. 

For quality assurance, 35 NCGold samples were provided for learning, and a round of evaluations was conducted on 50 withheld samples. We manually reviewed the LLM-generated translations for validation, correcting minor character choice errors. The choice of 35 samples was informed by preliminary tests, where fewer examples often led the model to refuse responses. Since there is no strict upper limit for the number of NCGold examples (aside from token limits), incorporating more high-quality examples could further enhance translation performance.

\begin{table}[ht]
  \centering
  \begin{tabular}{ccc}
    \hline
    \textbf{Round} & \textbf{Data Sample} & \textbf{Average Length} \\
    \hline
    1 & 1-30   & 10.33 \\
    2 & 31-60  & 13.63 \\
    3 & 61-90  & 16.27 \\
    4 & 91-120 & 19.40  \\
    5 & 121-150 & 23.10  \\
    6 & 151-180 & 31.73 \\
    \hline
  \end{tabular}
  \caption{Average sentence length across data samples for different rounds of annotations.}
  \label{tbl:gpt-data-annotation-target}
\end{table}

The UD Chinese GSDSimp dataset \cite{ud_chinese_gsdsimp} was used to construct a set of new Chinese sentences for annotation by GPT-4-Turbo. First, we filtered out sentences containing out-of-dictionary words based on our Nüshu-Chinese dictionary, and then categorized the remaining sentences by ascending length. Table \ref{tbl:gpt-data-annotation-target} details the data samples annotated in each round, along with the average sentence length. Arabic numbers were also removed to avoid biasing the translation evaluations. 

Through iterative annotations using ChatGPT and manual corrections, we progressively expanded the translation process to include longer sentences, resulting in a more comprehensive set of Nüshu-Chinese translation instances. This iterative approach to sentence length enabled the generation of progressively longer translations, surpassing the limitations of the existing dataset. However, translation accuracy for longer sentences was slightly lower than for shorter ones, likely due to the fact that all seed sentences were derived from short, ancient Chinese poems.

\subsubsection{Generation Pipeline Setup}
\begin{figure}[t]
  \centering
  \includegraphics[width=\linewidth]{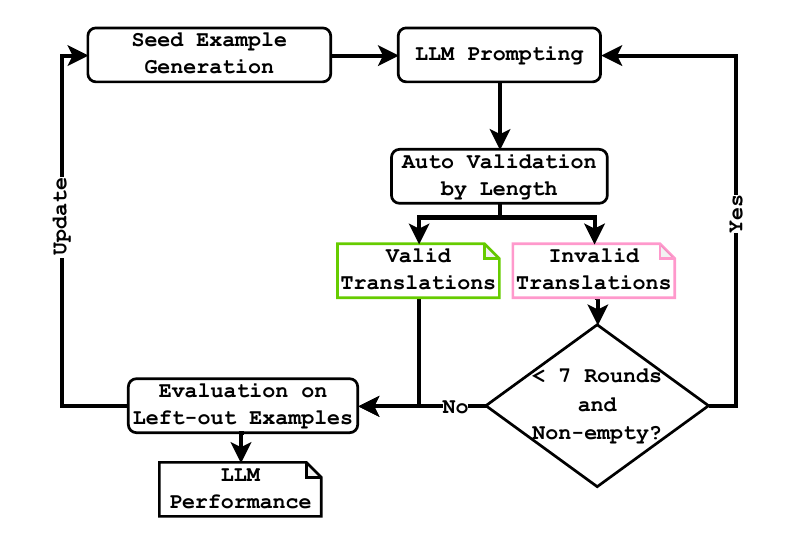}
  \caption{Pipeline for LLM-in-the-loop data annotation approach (Rounds 1-6).}
  \label{fig:gpt-data-annotation-pipeline}
\end{figure}

The pipeline of our LLM-in-the-loop data annotation approach is illustrated in Figure \ref{fig:gpt-data-annotation-pipeline}, where the LLM agent is queried to translate Chinese sentences into Nüshu using few-shot prompting. In the initial round (R1), seed translation examples are sampled from the NCGold dataset, while subsequent example sets are drawn from a combination of NCGold and NCSilver after the first round of translations. Beginning in the second round (R2), newly generated samples from the previous round are shuffled, with five samples selected to replace the top 5 samples in the original 35-sample list and the bottom five samples removed. For instance, in R2, the 35 training samples consisted of the top 5 new translations from R1, followed by 30 samples from the original set. By the third round (R3), the 35 samples included five new translations from R2, five from R1, and 25 from the original set. The control test consistently used the original 35 samples across all rounds. 

A rule-based translation length validator is integrated into the pipeline to ensure that each Nüshu character maps to a single Chinese character. This validator ensures that the original and translated contents have the same length. If a length mismatch occurs, the generation is rejected, and the LLM is prompted again to resolve the issue. The model is allowed up to 7 retries before moving on and classifying the instance as a failed translation. The enforced one-to-one character translations compensates for limited data, by providing additional linguistic instruction for Nüshu. Our experiments without enforced length validation reported a 31.37\% accuracy, versus 48.69\% with length validation, showing that data scarcity can be somewhat compensated through incorporating language-specific information.

\begin{figure}[t]
  \centering
  \includegraphics[width=\linewidth]{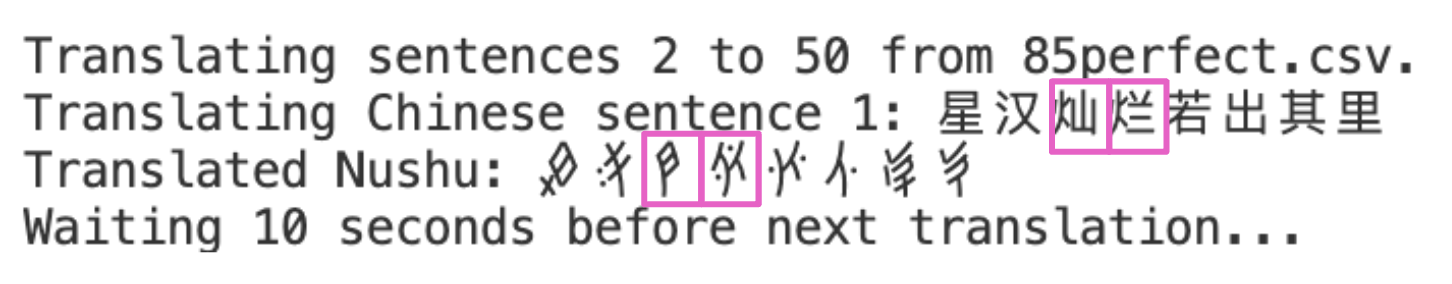}
  \caption{Translation of unseen characters (boxed in pink).}
  \label{fig:illustration:unseen}
\end{figure}

\begin{figure*}[t]
  \centering
  \includegraphics[width=\textwidth]{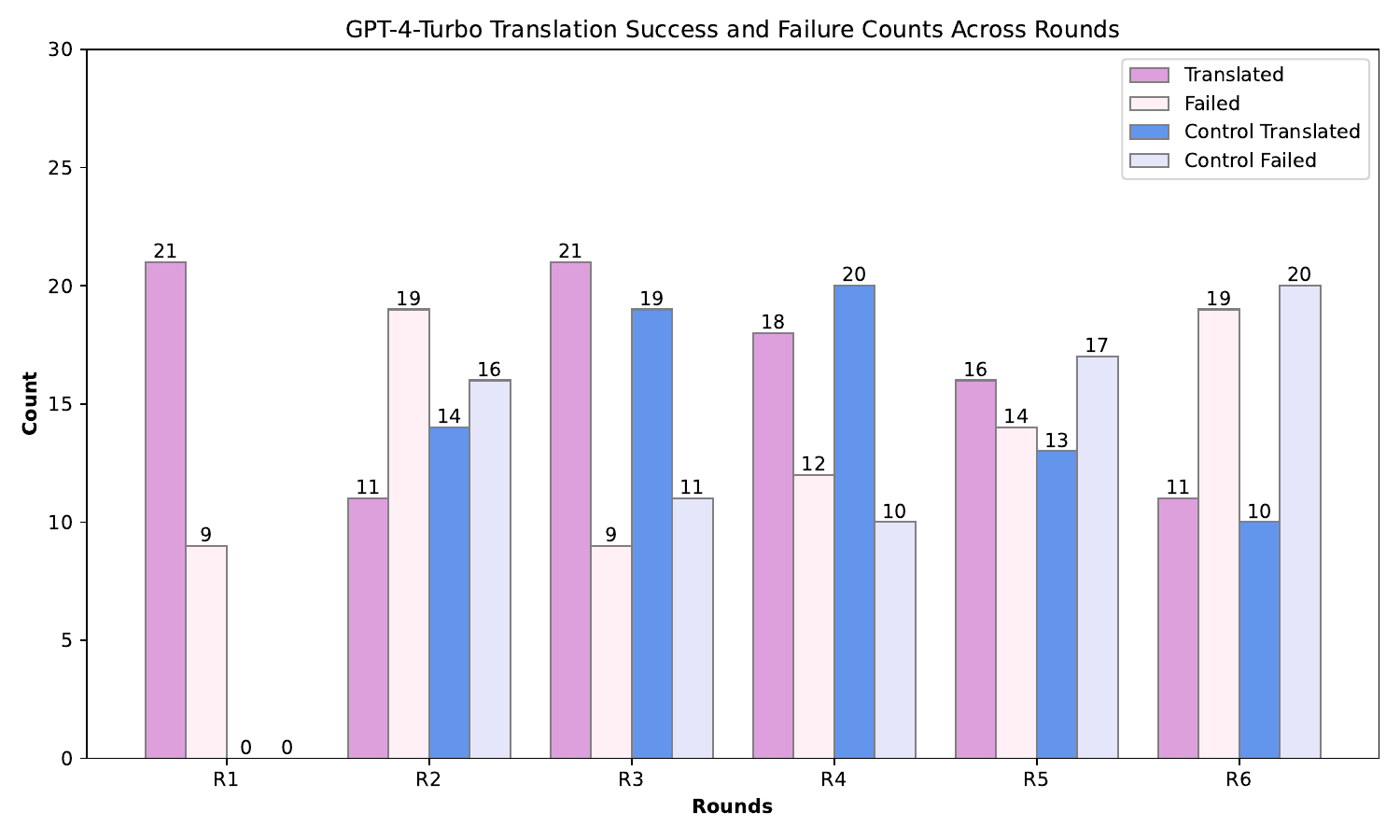}
  \caption{Translation success and failure across rounds.}
  \label{fig:illustration:bargraph}
\end{figure*}

It is critical to note that while the ChatGPT model is capable of generalizing knowledge from NCGold to unseen characters to produce seemingly valid translations, as exemplified in Figure \ref{fig:illustration:unseen}, expert knowledge of the Nüshu language is essential to validate these translations and register new characters. Due to the absence of such expert-level knowledge in our current work, we limit our testing to the model’s ability to translate only in-dictionary characters. Nevertheless, these findings suggest that our proposed framework could be highly applicable in future Linguistics research, aiding in the expansion of the Nüshu dictionary and the revival of the language.

\subsubsection{Data Annotation Outcomes}
Using only 35 sentences from NCGold as seed examples, the GPT-4-Turbo model achieved an accuracy of 48.69\% in perfectly translating the 50 left-out NCGold examples on a strict character-by-character match. A manual review of the translations revealed that minor translation errors were predominantly due to ambiguity in Nüshu character selection, as each Nüshu character can correspond to multiple Chinese characters and vice versa. 

\begin{figure}[t]
  \centering
  \includegraphics[width=\linewidth]{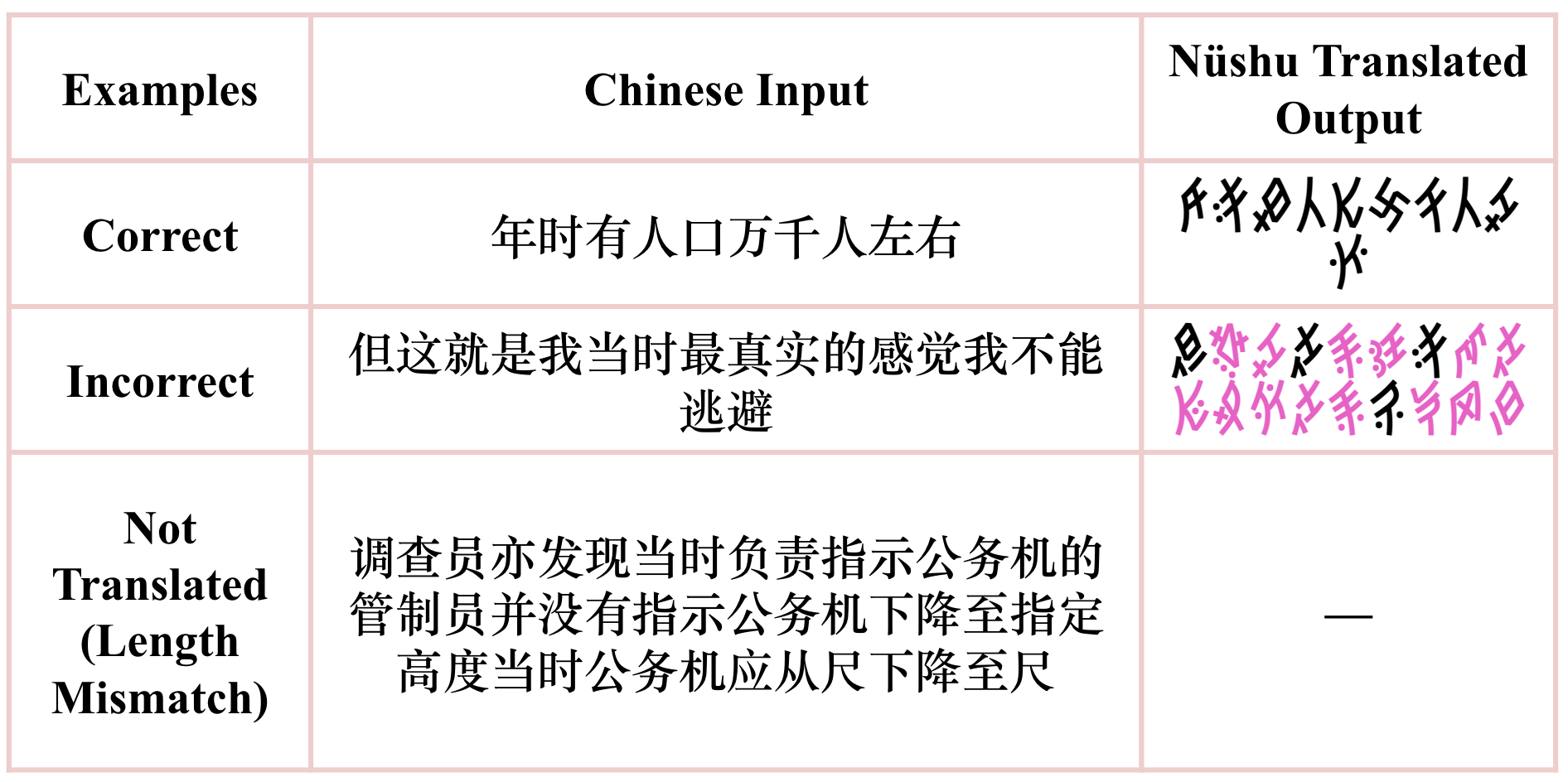}
  \caption{Out-of-domain examples; for the incorrect example, wrongly translated words are in pink.}
  \label{fig:gpt-data-annotation-examples}
\end{figure}

This complexity in mapping contributed to most of the discrepancies observed. For out-of-domain Chinese sentences, the GPT-4-Turbo model successfully translated 70.00\% of sentences with lengths below 16 characters. However, it frequently refused to respond when the sentence lengths exceeded 31 characters. Figure \ref{fig:illustration:bargraph} displays the number of successfully translated sentences and failed translations for each round, along with the results from the control group, which is always provided the original 35 learning samples. Figure \ref{fig:gpt-data-annotation-examples} exemplifies the correct, incorrect, and untranslatable responses from ChatGPT in our experiments.

\begin{table*}
  \centering
  \begin{tabular}{llllllll}
    \hline
    \textbf{Train Data}           
    & \textbf{BLEU-1} 
    & \textbf{BLEU-2} 
    & \textbf{BLEU-3} 
    & \textbf{METEOR} 
    & \textbf{ROUGE-1} 
    & \textbf{ROUGE-2} 
    & \textbf{ROUGE-L} 
    \\
    \hline
    100  & 0.0244 & 0.0074 & 0.0054 & 0.0125 & 0.0245 & 0.0    & 0.0245 \\
    200  & 0.0358 & 0.0147 & 0.0090 & 0.0220 & 0.0396 & 0.0048 & 0.0379 \\
    300  & 0.0351 & 0.0173 & 0.0133 & 0.0216 & 0.0381 & 0.0094 & 0.0381 \\
    400  & 0.0705 & 0.0234 & 0.0165 & 0.0398 & 0.0812 & 0.0036 & 0.0767 \\\hline
    421  & 0.0642 & 0.0231 & 0.0142 & 0.0372 & 0.0748 & 0.0065 & 0.0730 \\
    432  & 0.0893 & 0.0392 & 0.0313 & 0.0568 & 0.1053 & 0.0182 & 0.1037 \\
    453  & 0.0758 & 0.0394 & 0.0295 & 0.0542 & 0.0856 & 0.0230 & 0.0825 \\
    471  & 0.0977 & 0.0429 & 0.0296 & 0.0633 & 0.1174 & 0.0219 & 0.1115 \\
    487  & 0.1365 & 0.0685 & 0.0528 & 0.0942 & 0.1609 & 0.0408 & 0.1568 \\
    498  & 0.1093 & 0.0573 & 0.0404 & 0.0760 & 0.1367 & 0.0358 & 0.1308 \\
    \hline
  \end{tabular}
  \caption{\label{tbl:metrics}
    Evaluation results (averaged) for our Seq2Seq MT Model, evaluated on a fixed 100 test sample. 
  }
\end{table*}

\subsubsection{Other Low-Resource Languages}
To evaluate the scalability of our framework, we extended our investigation to additional low-resource languages with somewhat larger datasets than Nüshu but still underrepresented in mainstream language models like GPT-4-Turbo. Cherokee was selected as a representative Native American language, recognized for its moderate usage yet classified as low-resource \cite{zhang2022can}. Despite GPT-4-Turbo's partial familiarity with Cherokee \cite{shu2024transcending}, it struggled with accurate translations. Implementing our framework on a 14k Cherokee-English corpus \cite{zhang2020chren} yielded translation accuracies of 28\%, 40\%, and 42\% with 35, 150, and 300 sample sizes, respectively. These results highlight the persistent challenge of achieving higher accuracy in low-resource language translation without tailored linguistic adaptations.

\section{Applications}
Leveraging the NCGold and NCSilver datasets, we present both a Nüshu-Chinese language model and a Nüshu-to-Chinese translation model to demonstrate the potential of our data-annotation framework in advancing future research on Nüshu. Our experiments also underscore the critical need for further expanding the Nüshu corpus, enabling researchers to gain deeper insights into the language and the rich cultural heritage it embodies.

\subsection{Language Modeling in Nüshu} \label{sec:experiments:fasttext}
Having obtained a set of 598 Nüshu-Chinese translation sentence pairs, we employed the FastText library \cite{athiwaratkun2018probabilistic} to train a bilingual skip-gram language model \cite{mikolov2013efficient} in both Nüshu and Chinese. Specifically, we separated the Nüshu-Chinese translation pairs from NCGold and NCSilver into individual sentences, shuffled them, and utilized this bilingual corpus to train the FastText model. 
% TODO (IY): parameters for FastText training, e.g., epochs, window size, learning rate, vocab size, etc.

For the FastText model, we set the embedding vector size to 300, a window size of 10, and a minimum count of 5. We used the skip-gram model (sg=1), which is well-suited for smaller datasets and rare words, as it predicts surrounding words given a target word—particularly useful for the infrequent characters in the Nüshu corpus. 

The model was trained for 20 epochs to ensure convergence without over-fitting, with a negative sampling rate of 10 to speed up training by focusing on relevant examples.
The default learning rate of 0.05 is used for training our model.

\subsection{Nüshu-to-Chinese Translation} \label{sec:experiments:n2c}
We initially trained a Seq2Seq model for Nüshu-Chinese translation using training samples from NCGold, with a batch size of 3 and for 15 epochs. For the test set, we first shuffled the 500-sentence NCGold and fixed the first 100 samples as the test set. Despite the small dataset size, the model's performance was promising, comparable to models developed for other truly low-resource languages \cite{baruah2021low}. 

Encouraged by these results, we combined NCGold with NCSilver, resulting in a total of 598 sentences, to further improve translation quality. While ChatGPT demonstrated potential in translating Chinese text to Nüshu for NCSilver, it remains a closed-source model with limited interpretability, and its outputs are often noisy, requiring multiple queries for acceptable translations. To mitigate these issues and enhance both translation accuracy and model transparency, we retrained the Seq2Seq model using the combined NCGold and NCSilver datasets. We provide further details on this process in Section \ref{sec:discussion}.

\subsection{Nüshu Translates to Other Languages} \label{sec:experiments:n2o}
Given the limited availability of official Nüshu-Chinese translations, data for translating Nüshu into other languages is even more scarce. As mentioned in Section \ref{sec:related}, \citet{chiu2012heroines} published a collection of works featuring Nüshu calligraphy alongside English translations of their corresponding Chinese poems. However, this is not a direct Nüshu-to-English translation and lacks certain linguistic nuances or fidelity \cite{rosa2017theoretical}. Nevertheless, it introduces the right concept of using Nüshu as a bridge for multilingual translations. 

Our framework enables the construction of larger Nüshu datasets than ever before, making it possible to scale up Nüshu-to-other-language translation datasets when combined with such works. This approach opens up new avenues for expanding Nüshu’s reach across languages and enhancing the diversity of translation datasets.

\section{Discussion}% ~1-1.5 pages
\label{sec:discussion}
Given the promising early results of our Seq2Seq model trained on instances from NCGold against the fixed 100 test set, we hypothesized that the model's performance was limited by the small dataset size. We further proposed that, by continuing to use the data generation framework to incrementally enlarge the dataset, we could achieve even better model performance. To validate this, we trained the Seq2Seq model incrementally with subsets of NCGold, starting with 100, 200, 300 sentences, and so on, up to 400 sentences. The results, as shown in Table \ref{tbl:metrics}, confirm our hypothesis, demonstrating that larger dataset sizes consistently improve performance across all evaluated metrics.

To further improve the model, we gradually incorporated the 98 newly generated translations from NCSilver, adding them sequentially in the same quantities as generated during each round of the data generation process. As expected, model performance improved as the dataset expanded, reinforcing the importance of a larger and more diverse training corpus for Nushu-Chinese translation tasks. This trend is clearly reflected in the steady increase in BLEU, METEOR, and ROUGE scores, demonstrating a positive correlation between dataset size and translation quality.

However, we observed that performance peaked at a training dataset of 487 sentences. This dataset includes 400 original sentences from our book translations and 87 sentences generated through Round 5 (R5). Interestingly, performance declined when data from Round 6 (R6) was added. We hypothesize that this decline is due to the sentences in R6 being significantly longer than those in the rest of the dataset. As shown in Table \ref{tbl:gpt-data-annotation-target}, the average sentence length for R6 is 31.73, compared to 23.1 for R5, while the maximum sentence length in the original 500-sentence dataset is only 18.

We hypothesize that the model struggles with excessively long sentences because it was exposed to primarily shorter training data. Seq2Seq models, particularly those using a fixed-length encoder-decoder architecture, tend to perform well on sentence structures and lengths similar to those seen during training. When presented with longer, more complex sentences, the model may struggle to effectively encode the entire sentence into a fixed representation, leading to loss of important information, truncation, or difficulty decoding the translation. Furthermore, attention mechanisms can become less effective with longer sequences, especially when the model has not been fine-tuned or adequately trained. Consequently, incorporating longer sentences without balancing the model's exposure to various sentence lengths likely contributed to the performance drop observed after R6.

\section{Conclusion and Future Work}
In this paper, we introduced NüshuRescue, a novel approach leveraging LLMs to expand text corpora in endangered languages, with Nüshu as a case study. Our method facilitates research on these languages while minimizing reliance on extensive human annotations. Findings show that NüshuRescue can generate high-quality translations from Chinese to Nüshu with only 35 seed pairs, demonstrating its effectiveness in producing accurate translations with minimal input. The model also generalizes well across different sentence lengths and text domains. Notably, NüshuRescue does not require the underlying LLM to have prior knowledge of the target language, making it adaptable to other low-resource languages.

In addition, we digitized 500 Nüshu-Chinese sentence pairs from ancient Chinese poetry, which were used to train both a FastText language model and a Seq2Seq Nüshu-to-Chinese MT model, in combination with the LLM-generated translations. This dataset, encompassing 80\% of the only expert-validated Nüshu resource available, is a vital step in preserving Nüshu from extinction. In the absence of other digital resources, our paper serves as a foundational contribution, introducing Nüshu to the NLP community for the first time and providing a data augmentation framework. Looking ahead, NüshuRescue has potential for adaptation for other under-resourced languages, fostering the expansion of corpora and enabling the development of larger, more sophisticated language models.

\section*{Limitations}
In this paper, we introduced NüshuRescue, a LLM-based framework designed to scale up text corpora in resource-scarce languages. While our experiments demonstrated its effectiveness and generalizability, we encountered limitations in significantly expanding the Nüshu corpus due to the restricted scope of the current Nüshu-Chinese dictionary. The only available expert-validated source of Nüshu content comes from ``A Compendium of Chinese Nüshu'' \cite{zhao1992nushu_compendium}, which contains only 374 Nüshu characters and 1453 corresponding Chinese characters.

Although NüshuRescue demonstrated a strong ability to generalize to out-of-vocabulary characters, the absence of expert validation\footnote{Nüshu no longer has any native speakers, as the last known native speaker passed in 2004.} prevents us from safely releasing the generated data. Consequently, we sampled only 180 sentences from the UD Chinese GSDSimp dataset \cite{ud_chinese_gsdsimp} dataset to construct the NCSilver dataset, ensuring that all included Chinese characters were within the existing Nüshu vocabulary. However, as future studies of Nüshu incorporate more characters, NüshuRescue can be easily adapted to further expand the Nüshu corpus.

\section*{Ethics Considerations}
In this paper, we exclusively utilized publicly available resources, including the ``A Compendium of Chinese Nüshu'' book \cite{zhao1992nushu_compendium} and the UD Chinese GSDSimp dataset \cite{ud_chinese_gsdsimp} for our experiments. We relied on off-the-shelf LLMs, such as GPT-4-Turbo, as the backbone model for NüshuRescue. Furthermore, we have thoroughly documented the prompts used to generate Nüshu translations from the LLMs, detailing how these prompts were iteratively refined to improve the quality of the outputs.

Given these considerations, we do not foresee any ethical concerns arising from our work. Our NCGold and NCSilver datasets, as well as related code, have been made publicly available through our GitHub (\url{https://github.com/ivoryayang/NushuRescue}), ensuring transparency and fostering further research in this area.

\newpage
\section*{Acknowledgments}
This research was funded in part by the Dartmouth Alumni Research Award. We extend our gratitude to the Dartmouth graduate alumni for their generous support and commitment to fostering academic inquiry.

\newpage
\bibliography{custom}

\cleardoublepage

\appendix

\section{Appendix Figures}
This section presents supplementary figures referenced in the main content, providing additional visual support for illustrating our work.
\renewcommand{\thefigure}{A\arabic{figure}}
\renewcommand{\thetable}{A\arabic{table}}
\setcounter{figure}{0} 
\setcounter{table}{0} 
\begin{figure*}[!h]
  \centering
  \includegraphics[width=\linewidth]{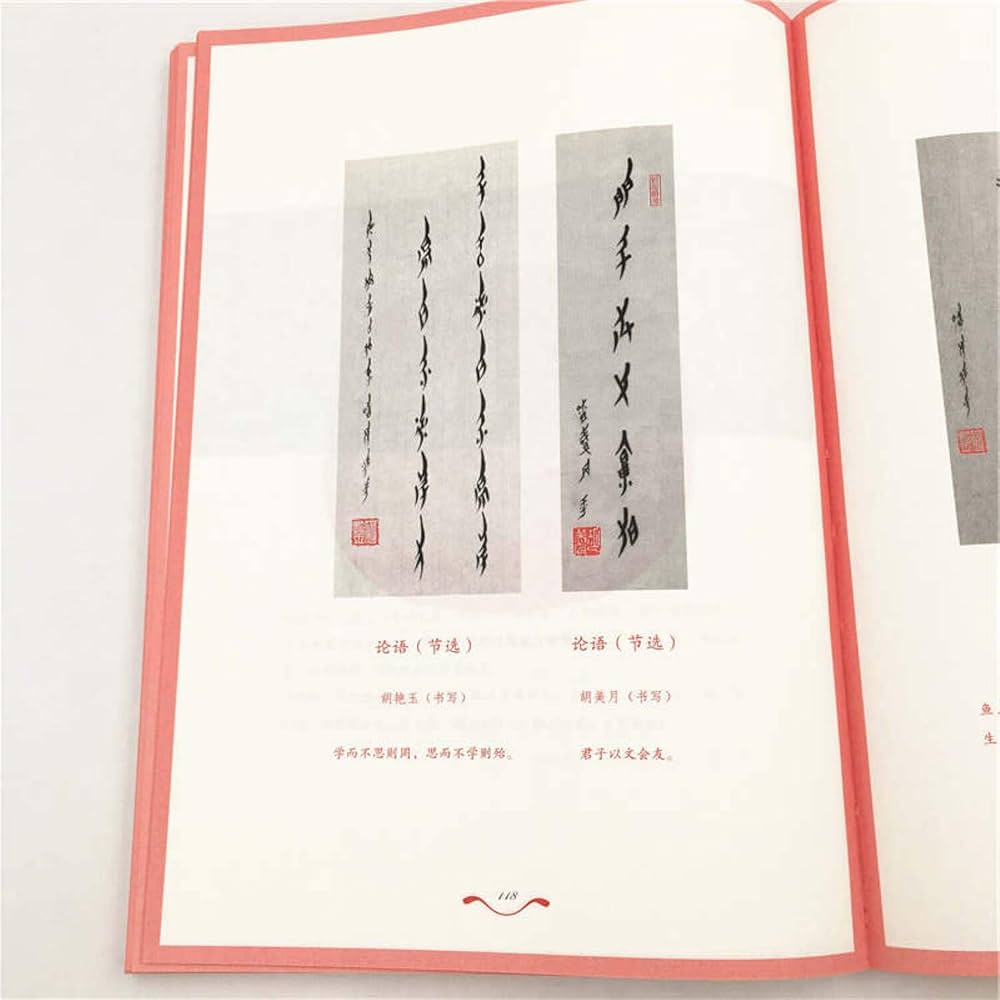}
  \caption{An example from the book (a public domain page scan included in the product description on Amazon, and therefore not subject to copyright restrictions).}
  \label{fig:appa}
\end{figure*}

\begin{figure*}[!h]
  \centering
  \includegraphics[width=0.5\linewidth]{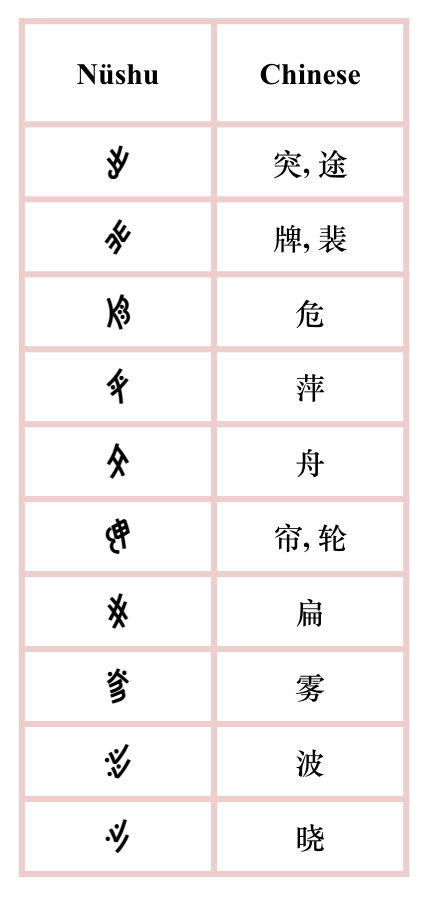}
  \caption{A small sample of newly mapped words not in the official Nüshu-Chinese dictionary; an original contribution.}
  \label{fig:appb}
\end{figure*}

\begin{figure*}[t]
  \centering
  \includegraphics[width=\linewidth]{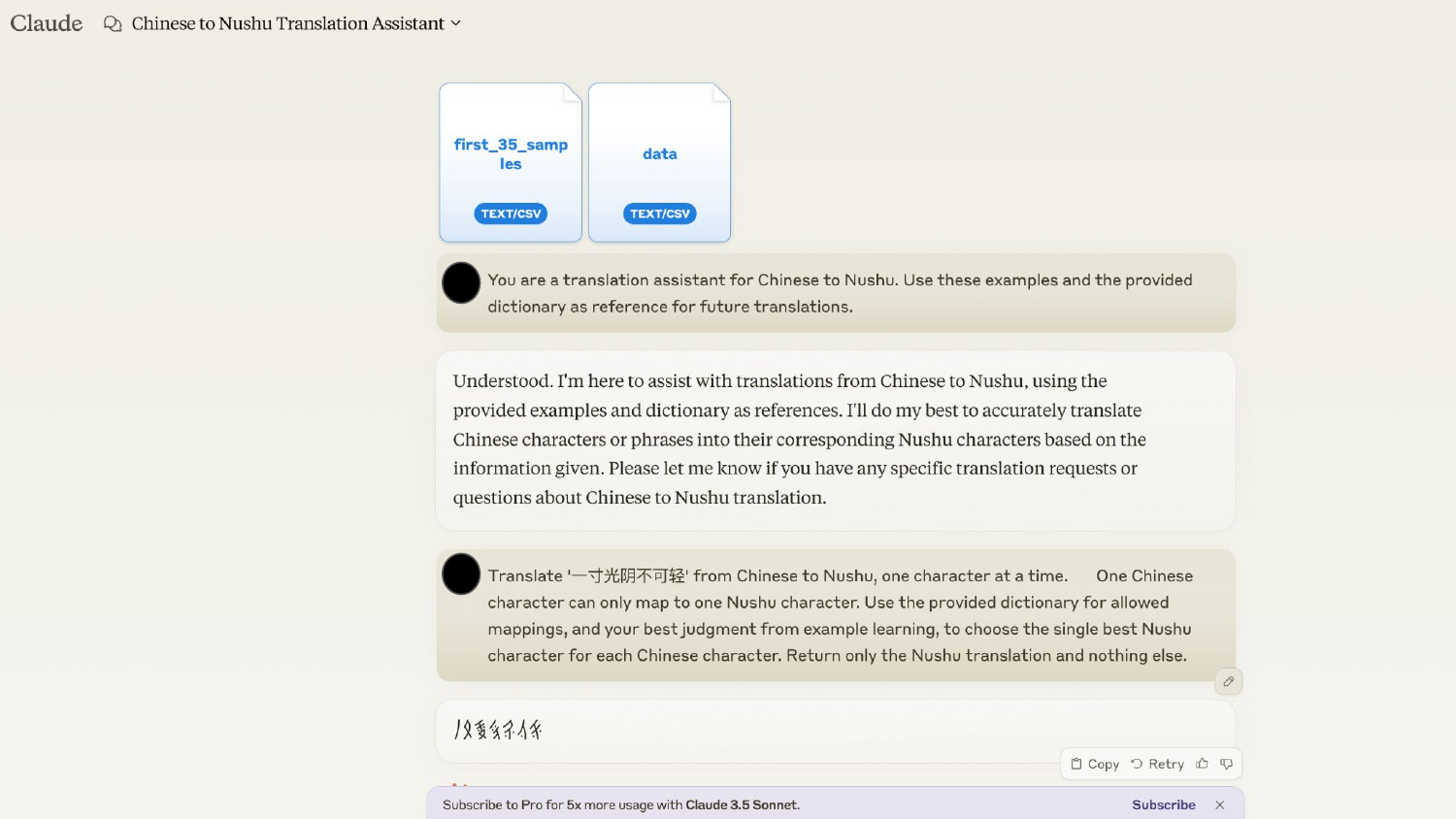}
  \caption{NüshuRescue is model-agnostic; it can work with Claude as well.}
  \label{fig:appc}
\end{figure*}

\end{document}